\documentclass{article}

\PassOptionsToPackage{numbers, compress}{natbib}

\usepackage[final]{neurips_2024}

\usepackage[utf8]{inputenc}

\usepackage[T1]{fontenc}    %
\usepackage{url}            %
\usepackage{booktabs}       %
\usepackage{amsfonts}       %
\usepackage{nicefrac}       %
\usepackage{microtype}      %
\usepackage{amsmath, amsthm, graphicx, amssymb}
\usepackage[dvipsnames,table]{xcolor}

\usepackage[noend]{algpseudocode}
\usepackage{algorithm,algpseudocode,tabularx}

\usepackage{multirow,bm,xfrac,pgf}
\usepackage{amsmath,graphicx,mathtools}
\usepackage{xspace,capt-of,float}
\usepackage[title]{appendix}
\usepackage{tabularx}
\usepackage{wrapfig, multirow,enumitem}
\usepackage{todonotes}%
\usepackage{bookmark}
\usepackage[capitalize]{cleveref}
\hypersetup{
    colorlinks,linktoc=all,pagebackref=true,
    linkcolor=BrickRed,
    citecolor=PineGreen,
    filecolor=Mulberry,
    urlcolor=NavyBlue,
    menucolor=BrickRed,
    runcolor=Mulberry,
}

\DeclareUnicodeCharacter{2212}{-}

\NewCommandCopy{\ORIcitep}{\citep}
\DeclareRobustCommand{\citep}{\leavevmode\unskip~\ORIcitep}

\NewCommandCopy{\ORIcitet}{\citet}
\DeclareRobustCommand{\citet}{\leavevmode\unskip~\ORIcitet}
\AtBeginEnvironment{appendices}{\crefalias{section}{appendix}}
\AtBeginEnvironment{appendices}{\crefalias{subsection}{appendix}}
\makeatletter
\DeclareRobustCommand\onedot{\futurelet\@let@token\@onedot}
\def\@onedot{\ifx\@let@token.\else.\null\fi\xspace}

\def\eg{\emph{e.g}\onedot,\xspace} 
\def\ie{\emph{i.e}\onedot,\xspace} 
 
 \def\vs{\emph{vs}\onedot}

\makeatother

\def\W{\ensuremath{\bm{W}}}

\newcommand{\R}{\mathbb{R}}

\everymath=\expandafter{\the\everymath\displaystyle}
\usepackage{array}
\newcolumntype{H}{>{\setbox0=\hbox\bgroup}c<{\egroup}@{}}

\title{Choice of PEFT Technique in Continual Learning: \\Prompt Tuning is Not All You Need}

\author{%
	Martin Wistuba \\
	AWS \\
	\texttt{marwistu@}
	\And
	Prabhu Teja Sivaprasad \\
	AWS \\
	\texttt{prbuteja@} 
	\And
	Lukas Balles\thanks{Work done at AWS} \\
	Aleph Alpha  
	\And
	Giovanni Zappella \\
	AWS \\
	\texttt{zappella@}
}

\begin{document}
\maketitle
\begin{abstract}
Recent Continual Learning (CL) methods have combined pretrained Transformers with prompt tuning, a
parameter-efficient fine-tuning (PEFT) technique.
We argue that the choice of prompt tuning
in prior works was an undefended and unablated decision, which has been uncritically adopted by
subsequent research, but warrants further research to understand its implications.
In this paper, we conduct this research and find that the choice of prompt tuning as a PEFT method hurts the overall performance of
the CL system. To illustrate this, we replace prompt tuning with LoRA in
two state-of-the-art continual learning methods: Learning to Prompt and S-Prompts. These variants
consistently achieve higher accuracy across a wide range of domain-incremental and class-incremental
benchmarks, while being competitive in inference speed. Our work highlights a crucial argument:
unexamined choices can hinder progress in the field, and rigorous ablations, such as the PEFT
method, are required to drive meaningful adoption of CL techniques in real-world applications.
\end{abstract}

\section{Introduction}
\label{sec:intro}

A practical need arising when successful models are deployed in real-world applications is the one
to update the model with new knowledge. Continual learning tackles this problem by studying the
incremental training of machine learning models on a sequence of datasets. The goal is to
efficiently learn from new data without forgetting knowledge obtained in the past. There is a
variety of approaches to continual learning. Replay-based
methods\citep{chaudhry2019tiny,buzzega2020dark} maintain a memory of previously-seen data and replay
it while training on the most recent data. Regularization
approaches\citep{kirkpatrick2017overcoming} penalize deviation of the model parameters from the
previously-found weight configuration.

Most of these traditional continual learning (CL) methods are architecture-agnostic and can be
applied to (pretrained) Transformer models\citep{vaswani2017attention} in a straight-forward
fashion, but either require to store data or have poor predictive performance.
Beyond that, there has been recent work on continual learning methods tailored specifically
to pretrained Transformer models using parameter-efficient fine-tuning method (PEFT) techniques from
NLP literature. %
Learning to Prompt \citep[L2P,][]{wang2022learning} is the first of these works and the authors
propose the use of prompt tuning\citep{lester2021power}. This work was the starting point for a
number of other works that have in common that they use prompt tuning as a PEFT method in continual
learning\citep{wang2022s,wang2022continual,dai2022lifelong,jung2023generating,smith2023coda,villa2023pivot,razdaibiedina2023progressive,khan2023introducing,liu2022incremental,gao2024consistent,roy2024convolutional}.
However, the reasons for this choice are unclear.

\subsubsection*{Revisiting PEFT Architecture Choices for CL}
\begin{figure}
    \renewcommand\sffamily{}
    \centering
    \input{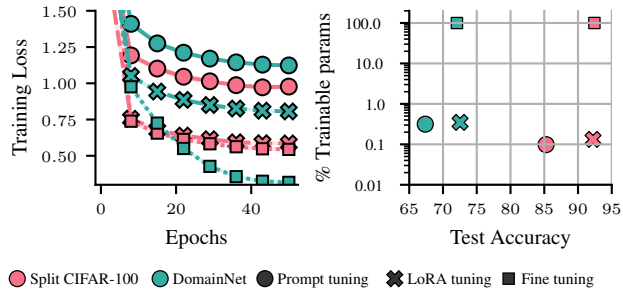}
    \caption{Is the choice of prompt tuning justified? We show train and test performance of a ViT-B/16
        trained on combined training set of all the datasets on Split CIFAR-100 and DomainNet and show the
        training loss dynamics on the left and the test performance on the right, for fine-tuning (training
        all parameters), prompt tuning (training prompt tokens and classifier layer), LoRA (training the low
        rank adapter and classifier layer). It is evident that prompt tuning converges to a higher loss
        (left), and performs poorly compared to LoRA (right), while being comparably parameter-efficient.
        Yet, a host of continual learning literature exists on using prompt tuning based techniques. We
        study if this design choice is justified. }
    \label{fig:selling_fig}
\end{figure}

Despite the prevalence of prompt tuning techniques in CL literature, it is undefended and unablated.
To understand this, consider a (non-continual) experiment as follows: We train
a ViT-B/16\citep{dosovitskiy2020vit} model on the combined training data of all domains for Split
CIFAR-100, and DomainNet. We plot the convergence characteristics of LoRA, prompt tuning, and
fine-tuning in \cref{fig:selling_fig}. We consider the cases of fine-tuning the full model, prompt
tuning, and training LoRA components of rank 1 for this combined dataset. As seen in
\cref{fig:selling_fig} (left), for both the datasets, fine-tuning always results in lower training
loss. This does not necessarily translate to better test performance as shown in
\cref{fig:selling_fig} (right), where LoRA and fine-tuning attain very similar performance, while
being vastly different on the number of parameters trained. It is also evident that, prompt tuning's
performance lags behind that of LoRA by a significant margin. With this evidence, we question the
utility of prompt tuning in the continual learning literature. LoRA is almost as parameter-efficient
as prompt tuning, and significantly more performant. Thus, the unablated choice of the PEFT method
can have an inordinate effect in the performance; say to the extent of being performant enough for
practical utility, or a paper being accepted at a reputed conference.

In this paper, we question the choice of using prompt tuning in CL by arguing that (some of) these methods do not intrinsically
use any specific properties of prompt tuning. We argue that any other PEFT methods could be used and
that this oftentimes improves  performance. In this way, we contribute to the literature that has
long studied the effects of improper evaluation of architectural choices. For example,
\Citet{Dodge2019ShowYW} show that several classical methods perform better than (then) modern deep
learning methods for NLP when considering hyperparameter choices. \citet{lucic2018Gans} come to
similar conclusions in the study of GANs, and \citet{rendle2019on} for recommender systems.
Continual learning literature has also a history of works pointing out brittle experimental
conditions. GDumb\citep{prabhu2020gdumb} showed that a na\"ive method that retrains networks from
scratch using a memory buffer for each task outperforms several advanced methods on a wide range of
continual learning scenarios. \citet{prabhu2023computationally} show that when considering the
computational requirements, the conclusions drawn about various famous CL methods do not hold.
\citet{janson2022a} challenge the very claim that the complex CL methods for pretrained models are
required at all. Without specific care for experimental setups (like hyperparameters, computational
requirements), the field is hindered from progressing beyond research into successful real-world
deployments.

In the rest of the paper we argue that our findings in \cref{fig:selling_fig} hold in the case of
continual learning too. For this, we decouple the PEFT method from the CL technique, and propose
LoRA variants of two aforementioned prompt-based CL methods, S-Prompts and L2P. Our experimental
evidence is in line with the inferences for the case of single dataset -- we empirically demonstrate
that these LoRA-based variants consistently achieve higher performance than their prompt-based
counterparts across a range of continual learning scenarios. At the same time, they stay competitive
in terms of the number of additional parameters as well as inference speed. Our findings do not only
improve upon the state-of-the-art, but also call into question the recent focus on prompt tuning for
continual learning and suggest a minor modification: in cases where the CL method is not
intrinsically related to the underlying PEFT architecture, LoRA should be preferred.
We are not aware of any method which has shown that prompt tuning is required nor are we are of any
method that could not be changed to use LoRA instead.

\section{Prerequisites and Prior Work}
\label{sec:prerequisites}

In this section, we will introduce concepts from continual learning as well as parameter-efficient
fine-tuning, which are relevant to the rest of the paper.

\subsection{Continual Learning}
Continual learning refers to the training of machine learning models on a sequence of datasets,
$\mathcal{D}_1, \mathcal{D}_2, \ldots$, which need to be processed sequentially. At episode $t$, we
would like our model to perform well on all the datasets observed up until that point.

Training from scratch, on $\mathcal{D}_{1:t} = \cup_{t^\prime=1}^t \mathcal{D}_{t^\prime}$ is
referred to as \emph{joint training} and is the gold standard in terms of performance. Continual
learning methods seek to achieve similar performance while avoiding the inefficiency of repeated
training from scratch. However, in this paper we consider the case where we have no access to older
datasets and can use only the most recent dataset to update a model.

We distinguish different continual learning \emph{scenarios}, depending on the relationship between
different datasets. In a class-incremental learning (CIL) scenario, each $\mathcal{D}_t$ introduces
previously-unseen classes. In contrast to that, a domain-incremental learning (DIL) scenario, the
set of labels is fixed, whereas the data distribution can change.

\subsection{Parameter-Efficient Fine-Tuning for Transformers}

Transformer architecture\citep{vaswani2017attention}, and its application to
vision\citep{dosovitskiy2020vit} have been widely used for their ability to learn powerful feature
representation. We present a brief overview of the Vision Transformer (ViT) architecture in
\cref{app:sec:vit}.

Due to their scale, fine-tuning a Transformer can be resource demanding. Parameter-efficient fine-tuning
(PEFT) methods have been devised in response. PEFT methods keep the bulk of a base model's
parameters frozen while fine-tuning a relatively small number of newly-added parameters (and/or a
subset of the base model's parameters) on a downstream task. This allows the adaptation of
large-scale models even on modest hardware. A host of PEFT methods exist with the first ones being
Adapter\citep{houlsby2019parameter}, where a small feed-forward module is inserted after each
multi-head attention layer. Here, we focus on prompt tuning and LoRA for their widespread adoption in the 
community.

\paragraph{Prompt Tuning}
Prompt tuning methods have evolved from the practice of prompting models with instructions prepended
to a given input. Instead of hand-crafted text prompts, prompt tuning \citep{lester2021power}
prepends trainable ``soft'' tokens in the input embedding space. That is, we allocate trainable
parameters $P\in \R^{L_\text{P} \times D}$, which are initialized randomly. For each input $x\in
    \R^{L_S \times D}$, we prepend $P$ and pass $[P; x]$ to the model. It increases the sequence length by
$L_\text{P}$, which affects the computational cost of the forward-backward pass through the model.
The trainable parameters for prompt tuning are $\{P, W^c, b^c\}$, a significant reduction from
fine-tuning.

\paragraph{Low-Rank Adaptation (LoRA)}
Low-rank adaptation \citep[LoRA;][]{hu2021lora} restricts the update of a weight matrix to a
low-rank subspace. For a given linear layer with a (frozen) weight matrix $W\in
    \R^{D_\text{in}\times D_\text{out}}$, LoRA adds a low-rank increment $BA$ with $B\in
    \R^{D_\text{in}\times r}$ and $A\in \R^{r \times D_\text{out}}$, where $r\ll \min(D_\text{in},
    D_\text{out})$. An input to the linear layer is then transformed as $z\mapsto Wz + BAz$. For
Transformer models, LoRA is usually applied to the query and value weight matrices of multi-head
attention layers, which are of size $D\times D$, where $D$ denotes the model's hidden dimension.
Hence, the trainable parameters are $\{A_l^Q, B_l^Q, A_l^V, B_l^V\}_{l=1}^L$ and $W^c, b^c$.

\section{Ablating the Choice of PEFT techniques}
\label{sec:current_work}

A majority of the works discussed in \cref{sec:intro} 
focused on techniques inspired from prompt
tuning. However, it is very unclear why that is the case. As we saw in \cref{fig:selling_fig}, LoRA
is substantially more performant than prompt tuning for a single downstream dataset. The question
remains whether this holds in the continual learning scenario.

To better investigate the points above, we propose LoRA-based variants of two popular
\emph{state-of-the-art} continual learning algorithms: S-Prompts\citep{wang2022s} and Learning to
Prompt\citep[L2P;][]{wang2022learning}. We chose S-Prompts for its simplicity, as it is
essentially a mixture of (parameter-efficient) experts. L2P, on the other hand, is prototypical for
a range of methods\citep{wang2022dualprompt,dai2022lifelong,smith2023coda} that maintain prompt
pools that are queried in an input-dependent fashion. Importantly, our proposed adaptations are
nearly surgical replacements, while inheriting the advantages of LoRA discussed before. A similar
analysis could be repeated with different PEFT methods but we leave that to future work. We describe
S-Prompts and L2P, and then propose our variants.

Let $x\in \R^{L_S \times D}$ a patchified and embedded input image. We denote the feature map of a
pretrained model, without an output layer, as $f\colon \R^{L\times D} \to \R^D$. It maps an input
sequence to a fixed-dimensional embedding space, \eg by selecting the embedding of a \texttt{[CLS]}
token or by averaging token embeddings across the sequence dimension. For simplicity, we assume that
the output dimension matches the input embedding dimension, which is the case for many contemporary
architectures.

\subsection{S-PEFT (S-X)}
\paragraph{S-Prompts}
S-Prompts\citep{wang2022s} learns a new set of prompts for each input dataset. Specifically, once a
new dataset $\mathcal{D}_t$ arrives, a new trainable prompt $P^{(t)} \in \R^{L_P\times D}$ as well
as a new output head $g_t$ are allocated and trained on $\mathcal{D}_t$ while keeping the base model
frozen. This leads to $t$ independent parameter-efficient expert models at time $t$. To identify the
prompt that is to be used at inference, a simple unsupervised method is used. For each input
dataset, $k$ prototype vectors are stored during training as follows: Each sample is embedded into a
feature space, and $k$-means clustering is used on that space to get the prototypes. In our
experiments, a ViT with \texttt{[CLS]} token features is used to embed the images. At inference, we
embed an image with the same network and choose the prompts corresponding to the closest prototypes.

\paragraph{S-LoRA}
As a simple mixture of experts strategy, S-Prompts is not intrinsically tied to prompt tuning. In
fact, it is straight-forward to use LoRA-based experts instead of prompt tuning. For each new
dataset $\mathcal{D}_t$, we allocate a new LoRA module as well as an output head $g_t$. Otherwise,
the training as well as the expert selection step remain unchanged from $S$-Prompts. We call this
strategy S-LoRA.

It is evident both S-Prompts and S-LoRA are specific instantiations of the larger class of
techniques based on PEFT methods, called S-PEFT (or S-X for short).

\subsection{Learning to PEFT (L2X)}
\paragraph{Learning to Prompt (L2P)}
S-Prompts allocates one set of prompt tokens per dataset. In contrast to that, Learning to Prompt
\citep[L2P;][]{wang2022learning} allocates a fixed-size \emph{pool} of prompts $\{P^{(1)}, \dotsc,
    P^{(M)}\} \subset \R^{L_P \times D}$ that is shared across datasets. Prompts are selected depending
on the input $x$ via a retrieval step. Specifically, each prompt is associated with a key vector
$k_i\in \R^{D_\text{out}}$ which lives in the embedding space of the base model and is initialized
randomly. A given input $x$ is mapped through the base model and a similarity score $s_i =
    \gamma(f(x), k_i)$ is computed for each key vector. The prompts corresponding to the $N$ highest
scores are selected, concatenated, and prepended to the input. Let $i_k(x) \in [M]$ be the index of
the $k^{\text{th}}$-highest score, then the prompted input is $x_\text{P} = [P^{(i_1(x))}; \ldots; P^{(i_N(x))};
    x]$. L2P uses a single shared classifier head $g$.

During training, we jointly train the prompt tokens, the key vectors, as well as the parameters of
the classifier head on the objective
\begin{equation}
    \label{eq:l2p_objective}
    \mathcal{L}(g(f(x_\text{P})), y) + \lambda \sum_{j=1}^N \gamma(f(x), k_{i_j(x)}).
\end{equation}
The first term is the standard loss whereas the second term encourages selected keys to move closer
to the corresponding query features.

The rationale of L2P is that some prompts will learn to specialize to specific datasets, whereas
others can encode shared knowledge. Since prompts are selected from the pool based on the
\emph{input}, no separate selection/identification step is needed at prediction time. For the same
reason, the method uses a shared classifier for all the datasets.

\paragraph{Learning to LoRA (L2L)}
For a LoRA-based variant of L2P, we allocate a pool of $M$ LoRA modules and use the exact same
key-based retrieval mechanism to select $N$ of them for a given input $x$. While prompts in L2P are
\emph{concatenated}, we combine LoRA modules in an \emph{additive} fashion. To make this precise,
let $W$ be a weight matrix and denote the selected corresponding LoRA matrices as $(A^{(i)},
    B^{(i)})$, $i\in [N]$. Then a given input $z$ is transformed as
\begin{equation}
    z \mapsto Wz + \sum_{i=1}^N B^{(i)}A^{(i)}z.
\end{equation}
Everything else is handled exactly as in L2P. We train with the loss function
\cref{eq:l2p_objective} (with non-prompted inputs) and the trainable parameters are the LoRA
modules, the key vectors, as well as the parameters of the output head. We will refer to this
strategy as L2L.

As with the S-X family, it is evident that both L2P and L2L are instantiations of the broader
Learning to PEFT (or L2X) class of methods.
\section{Experiments}\label{sec:experiments}

We now present experimental results comparing S-Prompts and L2P to their LoRA-based variants and other baselines.

\paragraph{Datasets:}

We show the performance of our proposed S-LoRA and L2L on both domain-incremental and
class-incremental benchmarks. For domain-incremental experiments, we run experiments with
CORe50\citep{lomonaco2017core50} and DomainNet\citep{peng2019moment}. For class-incremental
experiments, we use Split CIFAR-100\citep{zenke2017continual}, and Split Tiny
ImageNet\citep{tinyimagenet}. %
CORe50 is a benchmark for continual object recognition with 50 classes from 11 datasets with 8 of
them acting as the training set, and the rest as the test set. DomainNet is a benchmark for image
classification with 6 datasets each with 345 classes. Split CIFAR-100 and Split Tiny ImageNet refer to
splitting CIFAR-100\citep{krizhevsky2009learning} and Tiny ImageNet into 10 non-overlapping subsets.

\paragraph{Model: }To facilitate a fair comparison of baselines, we use a ViT-B-16 model\citep{dosovitskiy2020vit}
pretrained on ImageNet21k from the timm library (v0.6.5, Apache 2.0 license)\citep{rw2019timm,steiner2022how}.

\paragraph{Methods:}

We restrict our comparison to memory-free methods to ensure parity. In addition to L2P, L2L,
S-Prompts and S-LoRA, we add two more memory-free continual learning methods, EWC
\citep{kirkpatrick2017overcoming} and LwF \citep{li2017learning}. Our lower bound is the simple
fine-tuning baseline, which fine-tunes the model sequentially on each subsequent dataset. Finally,
we compare with three different variants of joint training, \ie training from scratch on all
previous data, combined with prompt tuning, LoRA and full fine-tuning. We implement all the methods %
in the open source library
Renate (v0.5.0, Apache 2.0 license) %
\citep{renate2023} with the LoRA-based
methods using the PEFT library (v0.5.0, Apache 2.0 license) \citep{peft}.
We run our baselines EWC and LwF using Avalanche (v0.4.0, MIT license)\citep{avalanche}.

\paragraph{Metrics}
We quantify the performance of all our experiments as the micro average of all the datasets. We
evaluate the final predictor $h_\ast\colon \mathcal{X}\to\mathcal{Y}$ that a method produces after
having observed all $T$ datasets. Let $\mathcal{D}^\text{te}_t$ denote a held-out test for dataset
$t$. The \emph{average accuracy} is defined as %
\begin{equation}
    \text{Average Accuracy} = \frac{1}{\sum_{t=1}^T\vert \mathcal{D}^\text{te}_t\vert} \sum_{t=1}^T \sum_{x, y \in \mathcal{D}^\text{te}_t} \mathbb{I}(h_\ast(x) = y).
    \label{eq:avg_acc}
\end{equation}

\paragraph{Hyperparameters}

We implement S-LoRA and L2L as drop-in replacements of their predecessors, S-Prompts and L2P,
respectively. This also implies sharing of hyperparameters whenever possible. Prompt-based and
LoRA-based methods differ in one key hyperparameter, which controls the number of trainable
parameters. Prompt-based methods have a prompt length $L_P$, whereas LoRA-based methods
choose a rank $r$. We present results with default choices for each, but also investigate the
behavior at various parameter levels. For L2P and L2L, we fix the pool size to $N=10$ and the number
of active modules to $M=5$. For S-Prompts and S-LoRA, we set the number of clusters to $k=5$ in the
domain-incremental settings following \citet{wang2022s}. For class-incremental settings, we set the
number to two times the number of new classes. We use a LoRA rank of $r=1$.

We use identical optimization hyperparameters too. For the S-X, we use the optimization
hyperparameters from \citet{wang2022s}. For L2P, we use the parameters prescribed by
\citep{wang2022learning}. However, for L2L we found that the default choice of Adam-W results in
very poor results, and thus we switch to SGD with momentum and keep the rest of the training
hyperparameters the same as L2P. The details of the hyperparameters are presented in
\cref{app:sec:hyperparams}.

\paragraph{Infrastructure and Compute Resources}
We ran our experiments using Amazon EC2 G5 Instances with a single NVIDIA A10G Tensor Core GPU (24 GB).
Obtaining our main results in \cref{tab:main_results} required 72 GPU days.
Additional compute was required for ablation studies and development.

\subsection{LoRA Variants Attain Higher Performance}\label{sub:lora-attains-higher-performance}

The main claim in this paper is that while prompt tuning is the predominant choice for continual
learning, there is no reason for why this is the case. In \cref{sec:current_work}, we have shown that there
is no technical reason for this choice. Now, in this section, we will demonstrate that prompt tuning
has no intrinsic properties that make it more suitable for continual learning. In order to do so, we
compare the prompt tuning variants against the LoRA variants, and summarize the results in
\cref{tab:main_results}. The performance improvements are apparent; across all four benchmarks, we
see that in the L2X family, L2L improves over L2P by on average about $6.1\%$, with the performance
on DomainNet increasing by approximately $10\%$. Similarly, in the S-X family, S-LoRA is better than
S-Prompts by about $4\%$ on average, and by $7.2\%$ on CORe50.

\definecolor{c1}{rgb}{0.6509803921568628, 0.807843137254902, 0.8901960784313725}
\definecolor{c2}{rgb}{0.12156862745098039, 0.47058823529411764, 0.7058823529411765}
\definecolor{c3}{rgb}{0.6980392156862745, 0.8745098039215686, 0.5411764705882353}
\definecolor{c4}{rgb}{0.2, 0.6274509803921569, 0.17254901960784313}
\begin{table}[ht]
    \caption{ Average accuracy and standard deviation for various continual learning scenarios.} %
    \label{tab:main_results}
    \small
    \begin{center}
        \begin{tabular}{lcHcccc}
            \toprule
            Method                                     & Buffer Size                                & CDDB                                       &
            CORe50                                     & DomainNet                                  & Split CIFAR-100                            & Tiny
            ImageNet                                                                                                                                                                                                                                                     \\
            \midrule
            Fine-Tuning                                & \multirow{7}{*}{$0$}                       & 66.14\scriptsize{$\pm7.44$}                &
            72.51\scriptsize{$\pm0.02$}                & 53.32\scriptsize{$\pm0.08$}                & 38.08\scriptsize{$\pm1.06$}                &
            50.78\scriptsize{$\pm1.28$}
            \\
            EWC~\cite{kirkpatrick2017overcoming}       &                                            & 62.67\scriptsize{$\pm7.44$}                &
            47.27\scriptsize{$\pm1.47$}                & 55.86\scriptsize{$\pm0.25$}                & 32.02\scriptsize{$\pm1.52$}                & 42.94\scriptsize{$\pm1.11$}
            \\
            LwF~\cite{li2017learning}                  &                                            & 75.07\scriptsize{$\pm3.31$}                &
            48.17\scriptsize{$\pm1.40$}                & 59.49\scriptsize{$\pm0.07$}                & 21.52\scriptsize{$\pm0.46$}                &
            23.16\scriptsize{$\pm1.60$}
            \\\rowcolor{c3}
            S-Prompts~\cite{wang2022s}                 & 0                                          & 77.80\scriptsize{$\pm0.19$}                &
            78.04\scriptsize{$\pm0.76$}                & 64.56\scriptsize{$\pm0.04$}                & 72.42\scriptsize{$\pm0.30$}                &
            63.38\scriptsize{$\pm0.19$}
            \\\rowcolor{c4}
            {\color{white}S-LoRA}                      &                                            & {\color{white}80.07\scriptsize{$\pm0.17$}} &
            {\color{white}85.23\scriptsize{$\pm0.21$}} & {\color{white}70.17\scriptsize{$\pm0.13$}} & {\color{white}73.07\scriptsize{$\pm0.16$}} &
            {\color{white}66.04\scriptsize{$\pm0.32$}}
            \\\rowcolor{c1}
            L2P~\cite{wang2022learning}                &                                            & 61.67\scriptsize{$\pm3.06$}                &
            90.34\scriptsize{$\pm0.63$}                & 47.91\scriptsize{$\pm0.35$}                & 79.42\scriptsize{$\pm1.90$}                &
            81.46\scriptsize{$\pm0.33$}
            \\\rowcolor{c2}
            {\color{white}L2L}                         &                                            & 58.86\scriptsize{$\pm2.97$}                &
            {\color{white}95.75\scriptsize{$\pm0.42$}} & {\color{white}57.87\scriptsize{$\pm0.12$}} & {\color{white}84.52\scriptsize{$\pm0.32$}} &
            {\color{white}85.59\scriptsize{$\pm0.18$}}
            \\
            \midrule
            Joint Prompt Tuning                        & \multirow{3}{*}{$\infty$}                  & 76.13\scriptsize{$\pm0.16$}                & 91.88\scriptsize{$\pm0.54$} & 67.38\scriptsize{$\pm0.04$} & 85.27\scriptsize{$\pm0.34$} & 79.38\scriptsize{$\pm0.27$} \\
            Joint LoRA                                 &                                            & 81.79\scriptsize{$\pm0.91$}                & 95.92\scriptsize{$\pm0.14$} & 72.53\scriptsize{$\pm0.08$} & 92.22\scriptsize{$\pm0.11$} & 89.36\scriptsize{$\pm0.28$} \\
            Joint Full Fine-Tuning                     &                                            & 90.47\scriptsize{$\pm0.29$}                & 91.55\scriptsize{$\pm0.28$} & 72.07\scriptsize{$\pm0.05$} & 92.43\scriptsize{$\pm0.11$} & 91.06\scriptsize{$\pm0.01$} \\
            \bottomrule
        \end{tabular}
    \end{center}
\end{table}

\Cref{tab:main_results} presents detailed results comparing several memory-free methods, and the
joint upper bounds described before. Given a pretrained model, fine-tuning which has been generally
considered a lower bound, performs better than some of the classic techniques like EWC and LWF on
multiple benchmarks. The S-X and L2X families outperform fine-tuning by a significant margin. It is
interesting to note that the joint training, which is often considered the upper bound, has a
varying performance based on the trainable parameters. Joint prompt tuning lags joint LoRA by at
least $4\%$ on CORe50, and about $10\%$ on Tiny ImageNet. Interestingly, joint fine-tuning is not
always the best performing and is often outperformed by joint LoRA.

In conclusion, if we are concerned with predictive performance, we observe no reason why prompt tuning should be preferred over LoRA. This confirms our main claim.

\subsection{Prompt-Based Methods Can Be More Run-Time Efficient}
\begin{figure*}[ht]
    \begin{minipage}{0.34\textwidth}
        \resizebox{\linewidth}{!}{
            \input{sec/figs/speed.pgf}
        }
        \caption{Measuring speed as a function of number of trainable parameters for Split CIFAR-100. We see
            that the prompt-based methods are faster only for a smaller number of trainable parameters.
        }
        \label{fig:speed_fig}
    \end{minipage}\hfill
    \begin{minipage}{0.63\textwidth}
        \resizebox{\linewidth}{!}{\input{sec/figs/performance_sweep_cifar100.pgf}}
        \caption{Performance of varying hyperparameters for Split CIFAR-100. We see that while
            increasing the number of trainable parameters does not improve necessarily result in an improved
            performance. For L2P, increasing the number of trainable parameters improves performance but does
            not reach the performance that L2L gets for a much fewer number of parameters. For the S-X family,
            it is apparent increasing number of parameters of S-LoRA is advantageous for performance, whereas
            S-LoRA performs poorer.}
        \label{fig:perf_sweeps}
    \end{minipage}
\end{figure*}

Prior works\citep{wang2022s} have compared different methods mostly in terms of additional parameters.
In the following, we consider a more relevant factor for practical usage: the run-time performance as
quantified by images processed per second. This is arguably a better metric than processing time per
image as it is better suited for batch parallelism in GPUs. This metric, however, is dependent on
the quality of the implementation. To
efficiently implement S-LoRA and L2L, we activate all the LoRA adapters available and mask the
outputs of the ones that are not relevant to a given input sample. This, while performing additional
computation, is amenable to a vectorized implementation in PyTorch\citep{paszke209pytorch}. While
this choice is not the optimal implementation, it is designed to maximize throughput when using
existing LoRA implementation from the PEFT library\citep{peft}. 
We benchmark our methods in two settings: best case and average
case. The best case setting assumes that all the samples in a batch belong to a single dataset and
that we are aware of the dataset ID using which we can select a single LoRA or prompt beforehand.
The average case for LoRA based methods assumes that each batch can have a mix of datasets.
Note that the average/best case
differentiation does not apply to L2X family as each sample is assigned its own adapter set.

In the S-X family, we vary only the prompt size for S-Prompts and the LoRA rank for S-LoRA. In the
L2X family, we vary the pool size $N$ and the prompt size for L2P, $N$ and rank $r$ for L2L. We show
the results of those hyperparameters in \cref{fig:speed_fig,fig:perf_sweeps}, and show the
throughput in images per second for various values of number of trainable parameters, and average
accuracy for number of trainable parameters, respectively. We plot a single line for each pool size
in \cref{fig:speed_fig} with different styles. We make several observations. The throughput is not a
monotonic function of the number of parameters; for the L2X the size of the pool influences the
number of parameters but not the throughput, which is dependent only on the selection size. This is
due to the specific choice of our implementation in L2L, where we compute the outputs at all LoRA
adapters and we discard the outputs for the inactive ones. Second, increasing the number of
parameters of the L2X family has little benefits. We see in \cref{fig:perf_sweeps}, that the
performance of the algorithms influences LoRA and prompt tuning differently. We see that increasing the
LoRA rank results in an increase in performance, where increasing the prompt size can possibly
worsen performance possibly due to overfitting. However, the performance of the LoRA adaptations is
always higher than the original prompt-based methods. The evidence from
\cref{fig:speed_fig,fig:perf_sweeps} makes it clear that the utility of prompt-based methods is
primarily run-time performance, especially when using smaller and fewer prompts, and LoRA based
techniques quickly outperform prompting techniques. Additionally, the run-time measurements can be
further improved through efficient LoRA implementations\citep{sheng2023slora,dettmers2023qlora}.

\subsection{S-X vs L2X: Factors Influencing Performance}\label{sub:s-x-vs-l2x}

It is apparent from \cref{tab:main_results}, there is no clear winner between the L2X and S-X
families. We investigate two factors in detail and improve upon these results in
\cref{app:sec:s-x-vs-l2x}. We summarize the important findings here.
\begin{itemize}[itemsep=0.0ex, left=-1ex, topsep=-3pt, leftmargin=3ex, itemindent=0.0ex]
    \item \textbf{Expert selection method:} The expert selection method, which determines the adapter to
          use for a given input, is more important for LoRA based methods (S-LoRA) than for S-Prompts on class-incremental problems. We study an improvement to the expert selection (called S-X++) which updates
          the feature extractor on the data from the first task, and observe that this improves the
          performance by $5\%$ on Split CIFAR-100 and $14\%$ absolute on Tiny ImageNet.
    \item \textbf{Sharing the classification layer:} S-X uses a classification head for each expert
          while L2X shares the classification head in all updates. For the class-incremental benchmarks, when
          using a shared head, we mask the logits for classes that are not present in the current dataset as
          it is done in L2X. We find this important for performance, an observation previously made in
          \citep{smith2023coda,ahn2021ss}. Whenever L2X shows better results, using a shared classifier leads
          to $7-8\%$ higher average accuracy. A shared classifier might be useful in cases where the input
          distribution changes little across datasets so as to efficiently share knowledge across datasets,
          and having independent classifiers is advantageous otherwise.
\end{itemize}

\section{Related Work}
\label{sec:related_work}
Using fine-tuning for continual learning, \ie training an existing model only on the most recent
data, is prone to catastrophic forgetting\citep{mccloskey1989catastrophic}. Early methods in
continual learning address this issue by adding regularization terms that penalize changing
important parameters\citep{kirkpatrick2017overcoming,li2017learning,zenke2017continual} or avoid
changing these parameters \citep{mallya2018piggyback,mallya2018packnet}. An alternative approach is
to continually add new parameters such that the new parameters learn new concepts without forgetting
old ones\citep{rusu2016progressive,fernando2017pathnet,li2019learn2grow}. Replay-based
methods\citep{chaudhry2019tiny,buzzega2020dark} make use of previously-seen data, however, we will
focus on memory-free methods only.

The massive success of Transformer models has inspired new model-growing methods for continual
learning which use pretrained Transformer models. The aforementioned S-Prompts\citep{wang2022s} and
L2P\citep{wang2022learning} will be discussed in detail in the next section. In particular, L2P has
spurred a host of follow-up work on prompt-based continual learning. \citet{wang2022continual} as
well as \citet{dai2022lifelong} split the prompt pool into dataset-invariant and dataset-specific
prompts. \citet{jung2023generating} introduce a trainable prompt generator model instead of
maintaining a fixed prompt pool, whereas \citet{smith2023coda} assemble prompts from learnable
prompt components. \citet{gao2024consistent} propose to train the classifier with prompts randomly
selected to align the training of prompts with testing, and \citet{roy2024convolutional} use a
hypernetwork to infer the prompts' parameters. \citet{razdaibiedina2023progressive} were inspired by
progressive neural networks and instead of using a set of fixed prompts, they simply keep adding
prompts which are all used at the same time. Inspired by language guidance,
\citet{khan2023introducing} add an additional loss terms which incentivize the model to align the
prompt keys to the task's language representation and map the same semantic space across each task.
While not equivalent to prompt tuning, \citet{douillard2022dytox} introduce dataset-specific special
tokens in the middle of an encoder-decoder architecture. \citet{villa2023pivot} devise a prompting
strategy specifically for class-incremental learning on video data. 
We make no claims to set new state-of-the-art results and we acknowledge that some of the very recent methods
achieve $1-2\%$ higher results than ours~\cite{gao2024consistent,roy2024convolutional}.
Our claim is that using LoRA instead of prompt tuning gives a significant improvement which will also
translate to these new prompt-based algorithms and in consequence further improve the state-of-the-art.

\section{Conclusions}\label{sec:conclusions}
In this paper, we investigate whether the usage of prompt tuning as PEFT method in continual learning algorithms is justified.
Our findings strongly suggest that it is not and we substantiate this claim through two key contributions.

First, we demonstrate that there is no technical challenge to replace prompt tuning with LoRA by deriving two LoRA-based variants of two popular continual learning methods: Learning to Prompt and S-Prompts. This exercise underscores the ease with which LoRA can be considered for existing and future continual learning algorithms.

Second, we provide empirical evidence that using LoRA improves predictive performance significantly by a large margin. Our comprehensive experiments across various datasets and scenarios consistently show the superior performance of LoRA-based models compared to their prompt tuning counterparts.
Furthermore, our analysis reveals that LoRA does not introduce any practically relevant overhead in terms of inference speed or memory footprint.
The recent work
on efficient LoRA implementations like quantized LoRA\citep{dettmers2023qlora} or methods for
concurrent LoRA adapters\citep{sheng2023slora} can further boost the run-time performance of LoRA-based CL algorithms whereas such improvements are not possible for prompt tuning-based variants.
Using the code that we will release with this paper, the community will be able to experiment with several more PEFT methods since our work leverage the widely used PEFT library from Huggingface.

Based on our findings, we conclude that there is no justification to use prompt tuning for all practical purposes and strongly recommend the adoption of LoRA over prompt tuning in continual learning algorithms.
We believe that this an important step towards closing the accuracy gap between continual learning and training from scratch and so in making continual learning a viable practical solution in real-world applications.

\section{Limitations}\label{sec:limitations}
We confine our experiments to using LoRA, and it can be extended using other
PEFT techniques like IA3\citep{liu2022few}, VeRA\citep{kopiczko2024vera}, DoRA\citep{liu2024dora}
which has been shown to be more parameter-efficient than LoRA, while having the same performance.
Furthermore, our run-time comparisons are dependent on the quality of the implementation itself;
further performance improvements need an overhaul of the underlying PEFT implementations, and is
beyond the scope of this work. To draw parallels to the original S-Prompts and L2P, we use the same
hyperparameters that were used in those works, but one may see additional benefits in modifying
those parameters; for example, applying LoRA to a subset of layers may result in the same performance with fewer parameters.

\section{Societal Impact}\label{sec:societal-impacts}
Continual learning can make training more energy-efficient since it reduces the need for extensive retraining and computational resources.
On the other hand, practitioners must be aware that these methods may inadvertently learn and retain information from their training data. This is a problem if the data contains sensitive or private information.

\newpage
\bibliographystyle{plainnat}
\bibliography{mybib}

\begin{appendices}
      \clearpage

\section{Vision Transformer}\label{app:sec:vit}
Vision Transformer (ViT)\citep{dosovitskiy2020vit}, inspired from the Transformers used for text, was
proposed as an alternative to convolutional nets for image understanding. It takes as input an image
$I\in \R^{W\times H\times 3}$, and extracts non-overlapping patches of size $P\times P$. Each patch
is embedded into a $D$ dimensional space. The resulting representation is of the shape
$\tfrac{WH}{P^2}\times D$. To this, a matrix of the same shape, called learned position encoding, is
added. A special token called the classification ([CLS]) token is concatenated. The output of this
is $X_0 \in \R^{L_S\times D}$ where $L_S = \tfrac{W\cdot H}{P^2} + 1$. This feature representation is
refined through $L$ layers of multi-head self attention (MHSA) layers.
\[\left.
    \begin{array}{ll}
        X^a_l & = \text{MHSA}(X_{l-1}) + X_{l-1} \\
        X_l   & = \text{FFN}( X^a_l) +  X^a_l
    \end{array}
    \right \} \forall l = 1 \dots L
\]

Each MHSA consists $H$ self attention (SA) modules in parallel. Each self attention module can be written as
\begin{equation}
    \text{SA}(X_l) = \mathrm{softmax}\left(\frac{X_lW^l_Q{W^l_K}^TX_l^T}{2\sqrt{d}}\right)X_l\W_V
\end{equation}
and the FFN as
\begin{equation}
    \text{FFN}(X_l) = \text{GeLU}(W^l_2\text{GeLU}(W^l_1X_l+b^l_1)+b^l_2).
\end{equation}
The [CLS] token at $X_L$ is used as the feature and is fed into a linear classifier $\R^{D}
    \rightarrow\R^{C}$ producing $C$ logits for classification. The trainable parameters for fine-tuning
are all the weights and biases $\{W^l_*, b^l_*\}_{l=1}^L$ and $\{ W^c, b^c\}$ of the classifier
layer.

\begin{figure*}[ht]
    \renewcommand\sffamily{}
    \resizebox{\textwidth}{!}{%
        \input{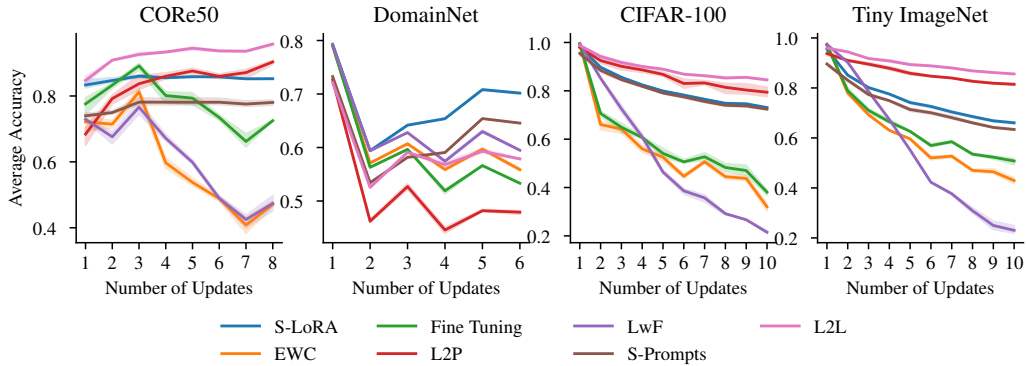}
    }
    \caption{We report the average accuracy obtained after each update. Ranking of LoRA \vs prompting-based methods does not change.}
    \label{fig:main_results_detailed}
\end{figure*}
\begin{table*}[t]
    \caption{Forgetting for all methods on the different benchmarks. Smaller means better.}
    \label{app:tab:forgetting}
    \begin{center}
        \begin{tabular}{lcccc}
            \toprule
            Method                               & CORe50                      & DomainNet                   & CIFAR-100                   & Tiny ImageNet               \\
            \midrule
            Fine Tuning                          & 9.34\scriptsize{$\pm1.11$}  & 20.25\scriptsize{$\pm0.17$} & 67.39\scriptsize{$\pm1.18$} & 51.36\scriptsize{$\pm1.41$} \\
            EWC~\cite{kirkpatrick2017overcoming} & 17.68\scriptsize{$\pm1.80$} & 19.01\scriptsize{$\pm0.17$} & 74.30\scriptsize{$\pm1.73$} & 60.20\scriptsize{$\pm1.29$} \\
            LwF~\cite{li2017learning}            & 17.34\scriptsize{$\pm2.59$} & 15.20\scriptsize{$\pm0.06$} & 85.83\scriptsize{$\pm0.53$} & 82.00\scriptsize{$\pm1.92$} \\
            S-Prompts~\cite{wang2022s}           & 0.23\scriptsize{$\pm0.27$}  & 1.63\scriptsize{$\pm0.02$}  & 8.41\scriptsize{$\pm0.12$}  & 10.18\scriptsize{$\pm0.20$} \\
            S-LoRA                               & 0.62\scriptsize{$\pm0.14$}  & 1.51\scriptsize{$\pm0.04$}  & 9.07\scriptsize{$\pm0.47$}  & 10.35\scriptsize{$\pm0.42$} \\
            L2P~\cite{wang2022learning}          & 0.00\scriptsize{$\pm0.00$}  & 21.54\scriptsize{$\pm0.46$} & 9.75\scriptsize{$\pm1.67$}  & 6.44\scriptsize{$\pm0.19$}  \\
            L2L                                  & 0.00\scriptsize{$\pm0.00$}  & 13.09\scriptsize{$\pm0.31$} & 7.34\scriptsize{$\pm0.31$}  & 4.96\scriptsize{$\pm0.35$}  \\
            \bottomrule
        \end{tabular}
    \end{center}
\end{table*}

\begin{table*}[t]
    \caption{Backward transfer for all methods on the different benchmarks. Larger means better.}
    \label{app:tab:backward_transfer}
    \small
    \begin{center}
        \begin{tabular}{lcccc}
            \toprule
            Method                               & CORe50                       & DomainNet                    & CIFAR-100                    & Tiny ImageNet                \\
            \midrule
            Fine Tuning                          & -5.95\scriptsize{$\pm0.80$}  & -20.25\scriptsize{$\pm0.17$} & -67.39\scriptsize{$\pm1.18$} & -51.36\scriptsize{$\pm1.41$} \\
            EWC~\cite{kirkpatrick2017overcoming} & -14.09\scriptsize{$\pm1.19$} & -19.01\scriptsize{$\pm0.17$} & -74.30\scriptsize{$\pm1.73$} & -60.20\scriptsize{$\pm1.29$} \\
            LwF~\cite{li2017learning}            & -14.82\scriptsize{$\pm2.45$} & -15.20\scriptsize{$\pm0.06$} & -85.83\scriptsize{$\pm0.53$} & -82.00\scriptsize{$\pm1.92$} \\
            S-Prompts~\cite{wang2022s}           & 1.05\scriptsize{$\pm0.44$}   & -1.63\scriptsize{$\pm0.02$}  & -8.41\scriptsize{$\pm0.12$}  & -10.18\scriptsize{$\pm0.20$} \\
            S-LoRA                               & 0.03\scriptsize{$\pm0.35$}   & -1.51\scriptsize{$\pm0.04$}  & -9.07\scriptsize{$\pm0.47$}  & -10.35\scriptsize{$\pm0.42$} \\
            L2P~\cite{wang2022learning}          & 7.77\scriptsize{$\pm0.31$}   & -21.54\scriptsize{$\pm0.46$} & -9.72\scriptsize{$\pm1.68$}  & -6.41\scriptsize{$\pm0.22$}  \\
            L2L                                  & 3.85\scriptsize{$\pm0.58$}   & -13.09\scriptsize{$\pm0.31$} & -7.31\scriptsize{$\pm0.31$}  & -4.96\scriptsize{$\pm0.35$}  \\
            \bottomrule
        \end{tabular}
    \end{center}
\end{table*}

\section{S-X vs L2X: Factors Influencing Performance}\label{app:sec:s-x-vs-l2x}
\begin{figure}[ht]
    \centering
    \resizebox{\linewidth}{!}{\includegraphics{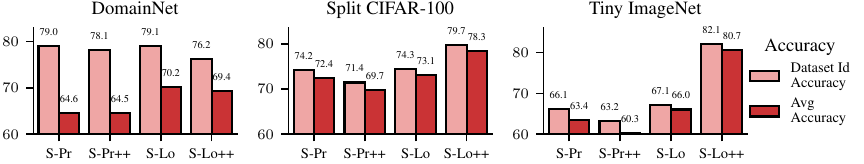}}
    \caption{S-Prompts (S-Pr) shows no positive change when using the prompts estimated for the first dataset
        to extract the feature representation (S-Pr++). However, using the LoRA modules of the first dataset (S-Lo++)
        to extract the features gives a big boost in identifying the right expert model and hence average
        accuracy.}
    \label{fig:plus_plus}
\end{figure}

From the results in \cref{tab:main_results}, we see that there emerges no clear winner between the L2X
and S-X families. While the LoRA variants always outperform prompt version in that family, L2X
dominates on CORe50, whereas the trend is reversed on DomainNet. L2X family is significantly better
than S-X on the class incremental benchmarks of Split CIFAR-100 and Tiny ImageNet. Here, we delve
into the possible reasons for this behavior. We chalk this out to two differences between the
methods.

\paragraph*{Factor 1 - Expert Selection Accuracy:}
Both S-LoRA and S-Prompts estimate one expert per dataset. %
Selecting the right one is crucial, in
particular in the class-incremental settings where selecting the wrong expert almost always results
in a wrong prediction due to the wrong classification head being chosen.
We use features from a pretrained model for the expert identification, and we
hypothesize the pretrained model's representation is not always sufficient to reliably identify the
expert. To test this hypothesis, we propose the following modification to the expert identification:
instead of using a pretrained model to extract features, we use the model after the first
adaptation. We term this modification S-X++. In \cref{fig:plus_plus}, we show the performance
changes from the original S-X variant to the S-X++ variants. We find that the modified S-Prompt++
always underperforms the original S-Prompts. On the other hand S-LoRA++ gets a big boost in
performance
for the class-incremental datasets (Split CIFAR and Tiny ImageNet). This
is due to the improved expert identification accuracy as seen in \cref{fig:plus_plus}. This,
however, does not improve the performance of S-LoRA to the levels of L2X.
\def\inc#1{\cellcolor{OliveGreen!10}#1}
\def\dec#1{\cellcolor{RubineRed!10}#1}

\begin{table}[ht]
    \caption{We report three different average accuracy metrics of different S-X variations:
        average accuracy (\cref{eq:avg_acc}) (overall) and average accuracy conditioned on whether the right expert
        was selected or not. }
    \label{tab:shared_classifier}
    \centering
    \small
    \begin{tabular}{@{}ll|cc|cc@{}}
        \toprule
        Dataset                          & Average Accuracy         & S-Prompts & S-Prompts-S & S-LoRA & S-LoRA-S \\ \midrule
        \multirow{3}{*}{DomainNet}       & Overall                  & 64.56     & {47.42}     & 70.17  & 59.03    \\
                                         & If right expert selected & 66.98     & {43.92}     & 72.40  & 56.78    \\
                                         & If wrong expert selected & 55.45     & {60.63}     & 61.71  & 67.48    \\\midrule
        \multirow{3}{*}{Split CIFAR 100} & Overall                  & 72.42     & {74.91}     & 73.07  & 80.59    \\
                                         & If right expert selected & 97.52     & {87.18}     & 98.20  & 91.04    \\
                                         & If wrong expert selected & 0.22      & {39.54}     & 0.23   & 50.28    \\ \midrule
        \multirow{3}{*}{Tiny ImageNet}   & Overall                  & 63.38     & {70.00}     & 66.04  & 82.98    \\
                                         & If right expert selected & 95.77     & {86.06}     & 98.39  & 93.39    \\
                                         & If wrong expert selected & 0.08      & 39.16       & 0.03   & 61.55    \\ \bottomrule
    \end{tabular}
\end{table}
\paragraph*{Factor 2 - Sharing the Classifier Is Important:}
A big difference between S-X and L2X is that S-X uses a classification head for each expert while
L2X shares the classification head in all updates. Here we ablate this architectural choice. L2X
does not support using multiple heads, and thus, this analysis focuses on S-X family. For the class-incremental benchmarks, when using a shared head, we mask the logits for classes that are not
present in the current dataset as it is done in L2X. We find this important for performance, an
observation previously made in \citep{smith2023coda,ahn2021ss}.

Results with a shared classifier in \cref{tab:shared_classifier} show its importance. Whenever L2X
shows better results, using a shared classifier leads to $7-8\%$ higher average accuracy. For
DomainNet, where L2X lags S-X, using a shared classifier leads to worse results. We posit that a
shared classifier is useful in cases where the input distribution changes little across datasets so
as to efficiently share knowledge across datasets, and having independent classifiers is
advantageous otherwise.

We combine the improved initialization and shared classifier and present the results in
\cref{app:tab:s_variants_results} in the Appendix. In a nutshell, while both these approaches seem
orthogonal, we observe no notable empirical gains from this combination.

\section{Hyperparameters}\label{app:sec:hyperparams}
In the following, we report the hyperparameters used for training and each individual update algorithm.

\subsection{Training Hyperparameters}

\subsubsection{L2X}
We use the hyperparameter used by \citet{wang2022learning} for the L2X methods. We have tried to use the same settings as for all other methods but they were significantly worse.

\begin{tabular}{p{0.5\linewidth}m{0.4\linewidth}}
    \toprule
    Hyperparameter & Value               \\ \midrule
    Epochs         & 5                   \\
    Optimizer      & L2P: Adam, L2L: SGD \\
    Learning rate  & 0.001875            \\
    Weight decay   & 0.0                 \\
    Batch size     & 16                  \\
    \bottomrule
\end{tabular}

\subsubsection{All Other Methods}
The following settings are inspired by the settings used by \citet{wang2022s}.

\begin{tabular}{m{0.5\linewidth}m{0.4\linewidth}}
    \toprule
    Hyperparameter          & Value                                                                          \\ \midrule
    Epochs                  & 50                                                                             \\
    Optimizer               & SGD                                                                            \\
    Learning rate           & 0.001                                                                          \\
    Momentum                & 0.9                                                                            \\
    Weight decay            & $2\cdot 10^{-4}$                                                               \\
    Learning rate scheduler & cosine annealing with $T_\text{max}=50$, $\eta_\text{min}=0$,  interval: epoch
    \\
    Batch size              & 128                                                                            \\
    \bottomrule
\end{tabular}

For CORe50 we use only 20 epochs and reduce $T_\text{max}$ to 20, accordingly. We increased the learning rate to 0.02.

\begin{figure*}
    \renewcommand\sffamily{}
    \resizebox{\textwidth}{!}{%
        \input{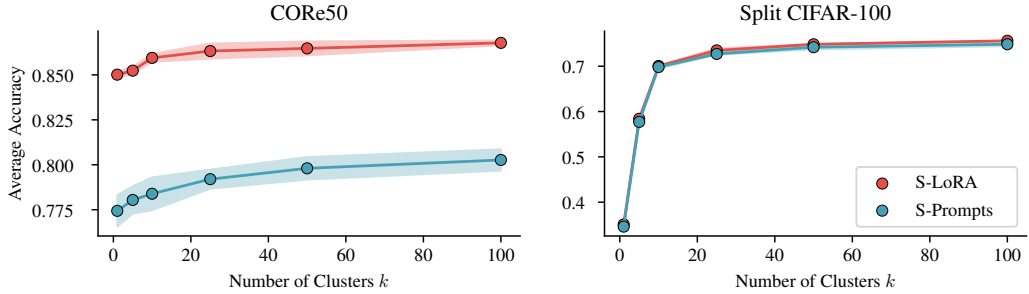}
    }
    \caption{Using more cluster centers has a positive impact on the expert selection accuracy and therefore improves average accuracy for both S-Prompts and S-LoRA.}
    \label{fig:cluster_ablation}
\end{figure*}

\begin{figure*}
    \renewcommand\sffamily{}
    \resizebox{\textwidth}{!}{%
        \input{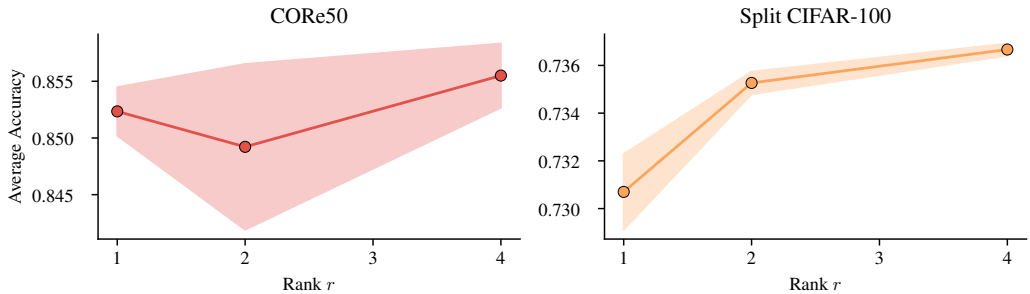}
    }
    \caption{Increase the rank for S-LoRA will also increase the number of additional parameters, but can improve the average accuracy.}
    \label{fig:rank_ablation}
\end{figure*}

\subsection{Method specific Hyperparameters}
We selected the hyperparameter settings used by the original authors.
We set S-LoRA and L2L accordingly to their prompt versions using a rank of 1.

\begin{tabular}{p{0.3\linewidth}m{0.6\linewidth}}
    \toprule
    Method    & Hyperparameters                     \\ \midrule
    EWC       & $\lambda=1$                         \\
    LwF       & $\lambda_o=1$, $T=2$                \\
    S-Prompts & $L_p=10$                            \\
    S-LoRA    & $r=1$                               \\
    L2P       & $\lambda=0.1$, $M=5$, $N=10$        \\
    L2L       & $\lambda=0.1$, $M=5$, $N=10$, $r=1$ \\
    \bottomrule
\end{tabular}

The S-X methods also have a hyperparameter $k$ that defines the number of dataset prototypes we save.
Following \citet{wang2022s}, we set $k=5$ for the domain-incremental scenarios CORe50 and DomainNet.
For class-incremental scenarios Split CIFAR-100 and Tiny ImageNet, we set $k$ to twice the number of new classes, i.e., 20 and 40, respectively.
We have ablated this choice in \cref{fig:cluster_ablation} which indicates small further improvements by further increasing $k$.

\section{More Insights in Our Experiments}
We use this section to report a couple more experiments to provide more insights in the behavior of the different methods.
We provide more detailed results and investigate the impact of some hyperparameters.

\subsection{Average Accuracy with Growing Number of Updates}
In the main paper, we reported the average accuracy after all model update steps.
With \cref{fig:main_results_detailed} we provide insights how the average accuracy changes with each update.
Ranking of the prompting-based methods \vs the LoRA-based methods is stable.
There are also no big changes in terms of the ranking for the other methods with the exception of LwF which does relatively well for the first updates.

\subsection{Forgetting and Backward Transfer}
We report the additional metrics forgetting and backward transfer in \cref{app:tab:forgetting,app:tab:backward_transfer}, respectively.
We define forgetting as
\begin{equation}
    \text{Forgetting} = \frac{1}{T-1}\sum_{t=1}^{T-1}\max_{t'\in\{t,\ldots,T\}}R_{t',t}-R_{T,t}
\end{equation}
and backward transfer as
\begin{equation}
    \text{Backward transfer} = \frac{1}{T-1}\sum_{t=1}^{T-1}R_{T,t}-R_{t,t}
\end{equation}
where $T$ is the total number of datasets considered in our experiment, i.e., the number of model updates, and $R_{i,j}$ is the model's test classification accuracy on dataset $\mathcal{D}_j$ after being sequentially trained on datasets $\mathcal{D}_1,\ldots,\mathcal{D}_i$.

L2L shows less forgetting and more backward transfer than L2P.
We notice that there is no forgetting on CORe50.
The reason for this is that the model is constantly improving on all datasets.
We observed a similar behavior when using the S-X variant with the shared classifier (S-X-S).

Technically, the S-X family does not suffer from forgetting nor benefits from backward transfer since the original model remains unchanged.
All changes in predictions for a specific dataset are caused by changes in the unsupervised expert selection.
This strategy is the same method for both and differences in the metrics are caused only by the different models.

\subsection{Ablating S-X Hyperparameters}
We provide additional experiments to investigate the importance of the number of clusters and the rank for S-X.
In \cref{fig:cluster_ablation} we visualize the impact of increasing number of clusters on the average accuracy of S-X.
It is no surprise that increasing the number of clusters has a positive impact on the expert selection accuracy and therefore the average accuracy.
We use $k=5$ for all domain-incremental settings (CORe50 and DomainNet) in the paper since it was suggested by \citet{wang2022s}.
We observe that this value is clearly non-optimal and higher values can be considered.
\citet{wang2022s} did not consider class-incremental scenarios in their work.
We decided against using $k=5$ for these cases and choose $k$ to be twice as high as the number of new classes seen per update.
The motivation behind this choice is that we need cluster centers that are able to represent the individual classes.
As can be seen in \cref{fig:cluster_ablation}, the default setting of $k=5$ does significantly worse on Split CIFAR-100 compared to $k=20$ which was selected in our experiments as the default.
The performance quickly grows until $k$ equals the number of new classes seen in each update (10) when it starts to flatten.

In our experiments, we always used a LoRA rank of $r=1$ for the only purpose that this adds the smallest possible number of new parameters.
However, such small values are typically not used, typical values are 16 or higher, and we expect larger ranks to perform better.
Given that parameter constraints are still important to us, we consider smaller ranks and report the results in \cref{fig:rank_ablation}.
We observe moderate gains on Split CIFAR-100, but no significant improvements on CORe50.

\subsection{S-X \vs L2X: Factors Influencing Performance}
In \cref{sub:s-x-vs-l2x}, we introduced two variations of S-X to investigate contributing factors that resulted in better empirical performance of L2X on some benchmarks.
We introduced S-X++, a variant that uses the pretrained model after the first update for feature extraction.
Furthermore, we introduced S-X-S which uses a common shared classification head instead of a single head per dataset.
We report all results including standard deviation in \cref{app:tab:s_variants_results}.
Furthermore, we report results for combining both variants referred to as S-X-S++.
For latter, we only report results for S-LoRA since S-Prompts++ did not improve over S-Prompts.
As mentioned in the main paper, combining both approaches adds no benefit.
While S-LoRA-S++ has a higher expert selection accuracy than S-LoRA-S, we notice that the average accuracy in cases where the wrong expert was selected decreases significantly.
It appears that while both variants appear orthogonal, they enable correct classification for the same otherwise wrongly classified instances.

\begin{table*}[t]
    \caption{ We report four different metrics of different S-X variations: expert selection accuracy,
        average accuracy (\cref{eq:avg_acc}), and average accuracy conditioned on whether the right expert
        was selected or not. }
    \label{app:tab:s_variants_results}
    \small
    \begin{center}
        \begin{tabular}{llccc|cccc}
            \toprule
            \multirow{2}{*}{Dataset}       & \multirow{2}{*}{Metric}                  &
            \multicolumn{3}{c}{S-Prompts}  & \multicolumn{4}{c}{S-LoRA}
            \\
                                           &                                          &       & ++    & S     &       &
            ++                             & S                                        & S++                                   \\
            \midrule
            \multirow{4}{*}{DomainNet}     & Expert Selection Accuracy                & 78.97
                                           & 78.11                                    & 79.03 & 79.10 & 76.25 & 78.95 & 76.27
            \\
                                           & Average Accuracy                         & 64.56 & 64.52 & 47.42 & 70.17 &
            69.38                          & 59.03                                    & 58.88                                 \\
                                           & Average Accuracy (right expert selected) & 66.98 & 66.96 & 43.92 & 72.40 &
            72.07                          & 56.78                                    & 56.34
            \\
                                           & Average Accuracy (wrong expert selected) & 55.45 & 55.83 & 60.63 & 61.71 &
            60.75                          & 67.48                                    & 67.05
            \\
            \midrule{}
            \multirow{4}{*}{CIFAR-100}     & Expert Selection Accuracy                & 74.21
                                           & 71.43                                    & 74.23 & 74.35 & 79.65 & 74.36 & 79.74
            \\
                                           & Average Accuracy                         & 72.42 & 69.67 & 74.91 & 73.07 &
            78.33                          & 80.59                                    & 80.67                                 \\
                                           & Average Accuracy (right expert selected) & 97.52 & 97.45 & 87.18 & 98.20 &
            98.31                          & 91.04                                    & 90.41
            \\
                                           & Average Accuracy (wrong expert selected) & 0.22  & 0.23  & 39.54 & 0.23  &
            0.08                           & 50.28                                    & 42.34                                 \\
            \midrule{}
            \multirow{4}{*}{Tiny ImageNet} & Expert Selection Accuracy                & 66.15
                                           & 63.17                                    & 65.76 & 67.11 & 82.08 & 67.30
                                           & 82.04                                                                            \\
                                           & Average Accuracy                         & 63.38 & 60.32 & 70.00 & 66.04 &
            80.67                          & 82.98                                    & 82.88                                 \\
                                           & Average Accuracy (right expert selected) & 95.77 & 95.41 & 86.06 & 98.39 &
            98.27                          & 93.39                                    & 92.98
            \\
                                           & Average Accuracy (wrong expert selected) & 0.08  & 0.14  & 39.16 & 0.03  &
            0.06                           & 61.55                                    & 36.73                                 \\
            \bottomrule
        \end{tabular}
    \end{center}
\end{table*}

\end{appendices}

\end{document}